\documentclass[review]{elsarticle}
\usepackage[caption=false,font=normalsize,labelfont=sf,textfont=sf]{subfig}%
\usepackage{lineno,hyperref}
\usepackage{color}%
\modulolinenumbers[5]
\journal{Journal of \LaTeX\ Templates}

\begin{document}
\begin{frontmatter}

%\begin{flushleft}
% {\conf{\large\textcolor{red}{Accepted by Pattern Recognition on 01 November 2022.}}}
% \end{flushleft}

\title{A Tri-attention Fusion Guided Multi-modal Segmentation Network\tnoteref{mytitlenote}}
\tnotetext[mytitlenote]{Accepted by Pattern Recognition on November 01, 2021.}

\author[mymainaddress,mysecondaddress,mythirdaddress]{Tongxue Zhou}
\author[mymainaddress,mythirdaddress]{Su Ruan\corref{mycorrespondingauthor}}
\cortext[mycorrespondingauthor]{Corresponding author}
\ead{su.ruan@univ-rouen.fr}
\author[mymainaddress,myfourthaddress]{Pierre Vera}
\author[mysecondaddress,mythirdaddress]{St\'{e}phane Canu} 

\address[mymainaddress]{Université de Rouen Normandie, LITIS - QuantIF, Rouen 76183, France}

\address[mysecondaddress]{INSA de Rouen, LITIS -Apprentissage, Rouen 76800, France}

\address[mythirdaddress]{Normandie Univ, INSA Rouen, UNIROUEN, UNIHAVRE, LITIS, France}

\address[myfourthaddress]{Department of Nuclear Medicine, Henri Becquerel Cancer Center, Rouen, 76038, France}

\begin{abstract}
In the field of multimodal segmentation, the correlation between different modalities can be considered for improving the segmentation results. Considering the correlation between different MR modalities, in this paper, we propose a multi-modality segmentation network guided by a novel tri-attention fusion. Our network includes N model-independent encoding paths with N image sources, a tri-attention fusion block, a dual-attention fusion block, and a decoding path. The model independent encoding paths can capture modality-specific features from the N modalities. Considering that not all the features extracted from the encoders are useful for segmentation, we propose to use dual attention based fusion to re-weight the features along the modality and space paths, which can suppress less informative features and emphasize the useful ones for each modality at different positions. Since there exists a strong correlation between different modalities, based on the dual attention fusion block, we propose a correlation attention module to form the tri-attention fusion block. In the correlation attention module, a correlation description block is first used to learn the correlation between modalities and then a constraint based on the correlation is used to guide the network to learn the latent correlated features which are more relevant for segmentation. Finally, the obtained fused feature representation is projected by the decoder to obtain the segmentation results. Our experiment results tested on BraTS 2018 dataset for brain tumor segmentation demonstrate the effectiveness of our proposed method.
\end{abstract}

\begin{keyword}
Multi-modality fusion \sep Correlation \sep Brain tumor segmentation \sep Deep learning 
\end{keyword}
\end{frontmatter}
% \linenumbers

\section{Introduction}
\label{sec1}
Multimodal segmentation using a single model remains challenging due to the different image characteristics of different modalities. A key challenge is to exploit the latent correlation between modalities and to fuse the complementary information to improve the segmentation performance. In this paper, we proposed a method to exploit the multi-source correlation and apply it to brain tumor segmentation task.

Magnetic Resonance Imaging (MRI) is commonly used in radiology to diagnose brain tumors, it is a non-invasive and good soft tissue contrast imaging modality, which provides invaluable information about shape, size, and localization of brain tumors without exposing the patient to a high ionization radiation \cite{liang2000principles,bauer2013survey,drevelegas2010imaging}. The commonly used sequences are T1-weighted (T1), contrast-enhanced T1-weighted (T1c), T2-weighted (T2) and Fluid Attenuation Inversion Recovery (FLAIR) images. In this  work, we refer to these images of different sequences as modalities. Different modalities can provide complementary information to analyze different subregions of gliomas. For example, T2 and FLAIR highlight the tumor with peritumoral edema, designated whole tumor. T1 and T1c highlight the tumor without peritumoral edema, designated tumor core. An enhancing region of the tumor core with hyper-intensity can also be observed in T1c, designated enhancing tumor core. Therefore applying multi-modal images can reduce the information uncertainty and improve clinical diagnosis and segmentation accuracy. 

Inspired by a fact that, there is strong correlation between multi MR modalities, since the same scene (the same patient) is observed by different modalities \cite{lapuyade2017segmenting, zhou2020brain}. We propose a novel tri-attention fusion to guide 3D multi-modal brain tumor segmentation network. A preliminary conference version appeared at MICCAI 2020 \cite{zhou2020brain}, which focused on the multi-modal brain tumor segmentation with missing modality. This journal version extended previous work, and applied it on the full modality brain tumor segmentation. The main contributions in this paper are: 1) A novel correlation description block is introduced to discover the latent multi-source correlation between modalities. 2) A correlation constraint using KL divergence is proposed to aide the segmentation network to extract the correlated feature representation for a better segmentation. 3) A tri-attention fusion strategy is proposed to re-weight the feature representation along modality-attention, spatial-attention and correlation-attention paths. 4) The first 3D multimodal brain tumor segmentation network guided by tri-attention fusion is proposed.

The rest of the paper is organised as follows. Section \ref{sec2} reviews the relevant prior work, Section \ref{sec3} details our proposed method, Section \ref{sec4} describes the data used and implementation details, Section \ref{sec5} presents the experiment results, Section \ref{sec6} gives a further discussion about our method, and Section \ref{sec7} concludes our work.

\section{Related work}
\label{sec2}
\begin{figure}[htb]
\centering
\includegraphics[width=\columnwidth]{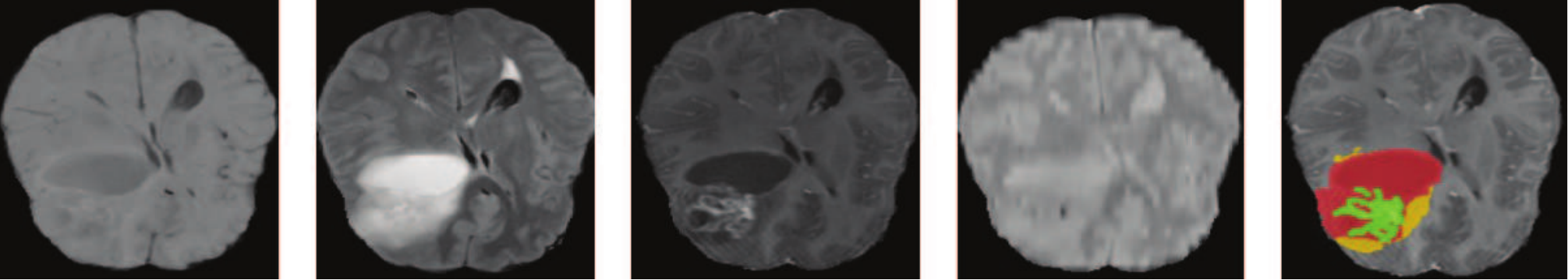}
\caption{Example of data from a training subject. The first four images from left to right show the MRI modalities: T1-weighted (T1), Fluid Attenuation Inversion Recovery (FLAIR), contrast enhanced T1-weighted (T1c), T2-weighted (T2) images, and the fifth image is the ground truth labels created by experts. The color is used to distinguish the different tumor regions: red: necrotic and non-enhancing tumor, yellow: edema, green: enhancing tumor, black: healthy tissue
and background.}
\label{fig1}
\end{figure}

A number of conventional brain tumor segmentation approaches haven been presented in recent years, including probability theory \cite{lapuyade2017segmenting}, kernel feature selection \cite{zhang2011kernel}, belief function \cite{lian2018joint}, random forests \cite{zikic2012decision}, conditional random fields \cite{yu2018semi} and support vector machines \cite{bauer2011fully}. However, the performance is limited due to the complex brain anatomy structure, different shape, texture of gliomas, and the low contrast of MR images (see \autoref{fig1}).

Recently, various deep learning-based approaches have been successfully designed for brain tumor segmentation, such as CNN \cite{wang2017automatic, dolz2020deep}, FCN \cite{zhao2018deep} and U-Net \cite{chen2020brain, myronenko20183d, wei2019m3net, chen2019dual, zhang2021cross}. Wang et al. \cite{wang2017automatic} developed a cascaded system to segment brain tumor using three binary segmentation subnetworks. Chen et al. \cite{chen2020brain} proposed a novel deep convolutional symmetric neural network, which combines the symmetry prior knowledge into brain tumor segmentation. Zhao et al. \cite{zhao2018deep} proposed a deep learning model integrating FCNNs and CRFs for brain tumor segmentation. Myronenko et al. \cite{myronenko20183d} proposed a segmentation network for brain tumor from multimodal 3D MRIs, where variational auto-encoder branch is added into the U-net to further regularize the decoder in the presence of limited training data. Wei et al. \cite{wei2019m3net} proposed a multi-model, multi-size and multi-view deep model for brain tumor segmentation. Dolz et al. \cite{dolz2020deep} presented an ensemble of deep CNNs to segment isointense infant brains in multi-modal MRI images. Chen et al. \cite{chen2019dual} proposed a dual-force training strategy to explicitly encourage deep models to learn high-quality multi-level features for brain tumor segmentation. Zhang et al. \cite{zhang2021cross} proposed a novel cross-modality deep feature learning framework for brain tumor segmentation, which consists of a cross-modality feature transition process and a cross-modality feature fusion process. Pinto et al. \cite{pinto2018hierarchical} introduced an automatic hierarchical brain tumour segmentation pipeline using Extremely Randomized Trees with appearance- and context-based features.

For multi-modal segmentation task, exploiting the complimentary information from different modalities plays an essential role in the final segmentation accuracy. The single-encoder-based method and multi-encoder-based method are the common used network frameworks \cite{zhou2019review}. The single-encoder-based method \textcolor{blue}{\cite{wang2017automatic, kamnitsas2017efficient}} directly fuses the different multi-source images in the input space, while the correlations between different modalities are not well exploited. However, the multi-encoder-based method \cite{tseng2017joint}, %\cite{chen2018mmfnet}, \cite{valindria2018multi}
applied separate encoders to extract individual feature representations, respectively. And it can achieve a better segmentation result than the former one \cite{zhou2020multi}. Since the effective feature representation can attribute to a better segmentation performance. Inspired by the attention mechanism \cite{roy2018concurrent}, in this paper, we first proposed a dual attention based fusion block to selectively emphasize feature representations, which consists of a modality attention module and a spatial attention module. The proposed fusion block uses the individual features obtained from encoders to derive a modality-wise and a spatial-wise weight map that quantify the relative importance of each modality’s features and also of the different spatial locations in each modality. These fusion maps are then multiplied with the individual feature representations to obtain a fused feature representation of the complementary multi-modality information. In this way, we can discover the most relevant characteristics to aide the segmentation.

For multi-modal MR brain tumor segmentation, since the four MR modalities are from the same patient, there exists a strong correlation in the tumor regions between modalities \cite{lapuyade2017segmenting}. Therefore, we proposed a correlation attention module, it consists of a correlation description block and a KL divergence based correlation constraint. It can exploit and utilize the correlation between modalities to improve the segmentation performance. In the correlation attention module, a correlation description block is first used to exploit the correlation between the spatial-attention feature representations, and then a correlation constraint based on KL divergence is used to guide the segmentation network to learn the correlated features to enhance the segmentation result. The novelty of this method is capable of exploiting and utilizing the latent multi-source correlation to help the segmentation. The proposed method can be generalized to other applications.

\section{Multi-modal segmentation with correlation constraints}
\label{sec3}
\begin{figure*}[htb]
\centering
\includegraphics[width=\textwidth]{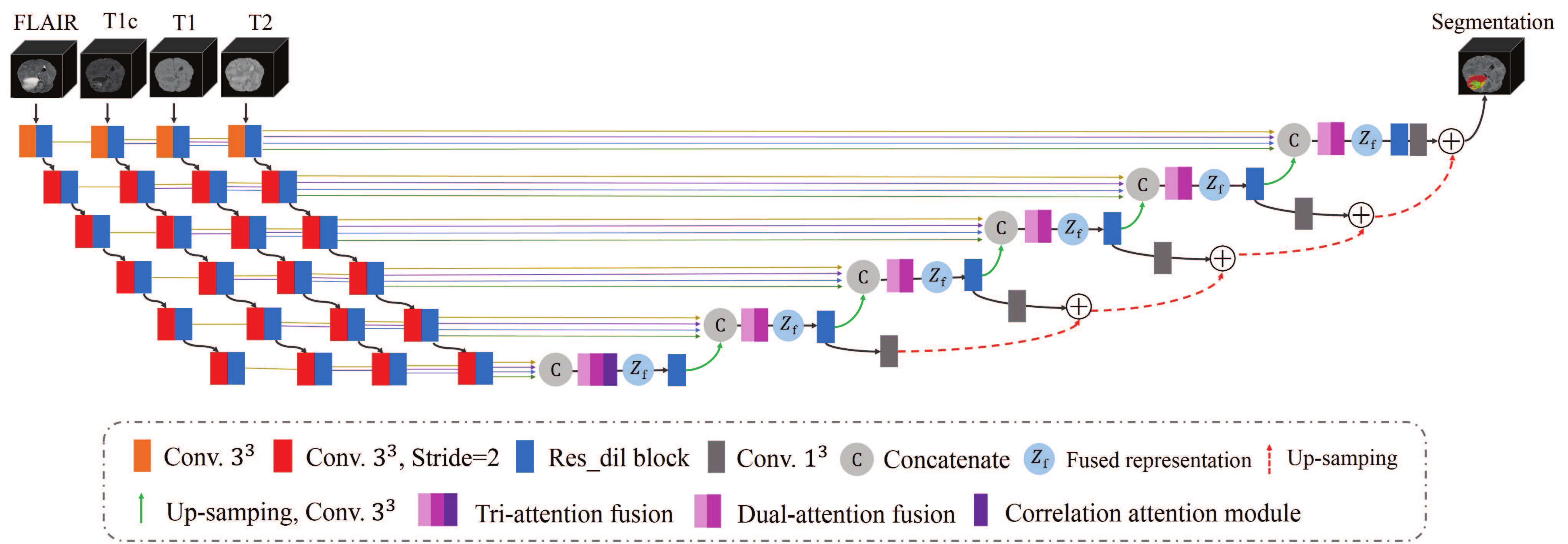}
\caption{Overview of our proposed segmentation network. The backbone is a multi-encoder based 3D U-Net, the separate encoders enable the network to extract the independent feature representations. The proposed dual-attention fusion block is to re-weight the feature representations along modality and space paths. The tri-attention fusion block consists of the dual-attention fusion and a correlation attention module.}
\label{fig2}
\end{figure*}

In this paper, we aim to exploit the multi-source correlation between modalities and utilize the correlation to constrain the network to learn more effective feature so as to improve the segmentation performance. U-Net is a neural network architecture widely used to medical image segmentation. The basic structure of a U-Net architecture consists of two paths. The encoder path is to extract feature representations at multiple different levels. The decoder path allows the network to project the discriminative features learnt by the encoder to the pixel space to get a dense classification. To learn complementary features and cross-modal inter-dependencies from multi-modality MRIs, we applied the multi-encoder based U-Net framework. It takes 3D MRI modality as input in each encoder. Each encoder can produce a modality-specific feature representation. At the lowest level of the network, the tri-attention fusion block is used, which includes a dual-attention fusion block and a correlation attention module. The dual-attention fusion block can re-weight the feature representation along modality-wise and spatial-wise. The correlation attention module is to first exploit the latent multi-source correlation between the spatial-attention feature representations. Then, it uses a correlation based constraint to guide the network to learn the effective feature information. Finally, the fused feature representation is projected by decoder to the label space to obtain the segmentation result. The overview of the proposed network is described in \autoref{fig2}.

\subsection{Encoder and Decoder}
\label{3-1}
It's likely to require different receptive fields when segmenting different regions in an image, a standard U-Net can’t get enough semantic features due to the limited receptive field. Inspired by dilated convolution, we use residual block with dilated convolutions (rate = 2, 4) (res\_dil block) on both encoder part and decoder part to obtain features at multiple scale. The encoder includes a convolutional block, a res\_dil block followed by skip connection. All convolutions are $3\times3\times3$. Each decoder level begins with up-sampling layer followed by a convolution to adjust the number of features. Then the upsampled features are combined with the features from the corresponding level of the encoder part using concatenation. After the concatenation, we use the res\_dil block to increase the receptive field. In addition, we employ deep supervision~\cite{isensee2017brain} for the segmentation decoder by integrating segmentation results from different levels to form the final network output.

\subsection{Tri-attention Fusion Strategy}
\label{3-2}
The purpose of fusion is to stand out the most important features from different source images to highlight regions that are greatly relevant to the target region. Since different MR modalities can identify different attributes of the target tumor. In addition, from the same MR modality, we can learn different content at different positions. Inspired by the attention mechanism \cite{roy2018concurrent}, we propose a dual-attention fusion block to enable a better integration of the complementary information between modalities, which consists of a modality attention module, and a spatial attention module.

\begin{figure}[!t]
\centering
\subfloat[]{\includegraphics[width=1.5in]{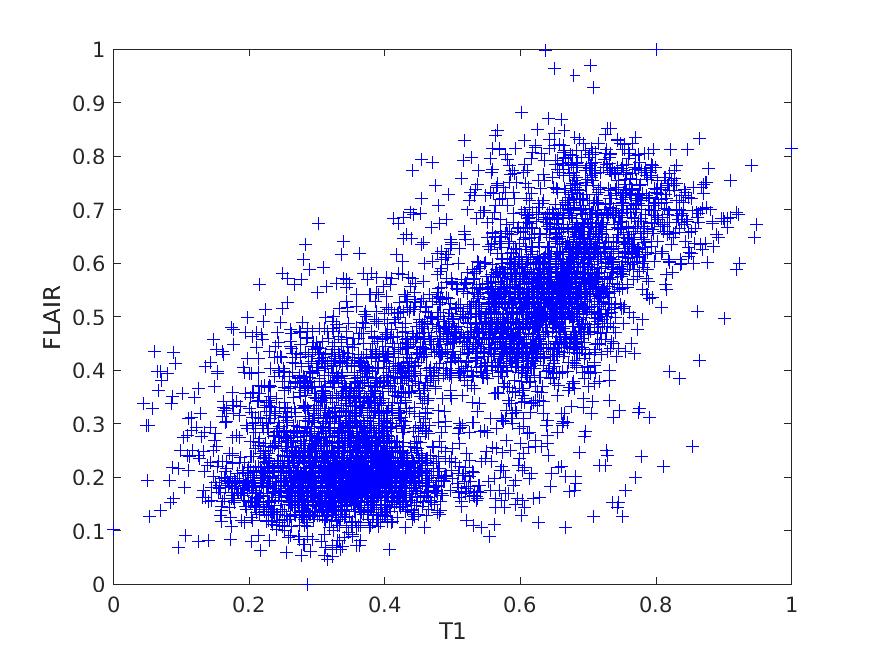}%
}
\hfil
\subfloat[]{\includegraphics[width=1.5in]{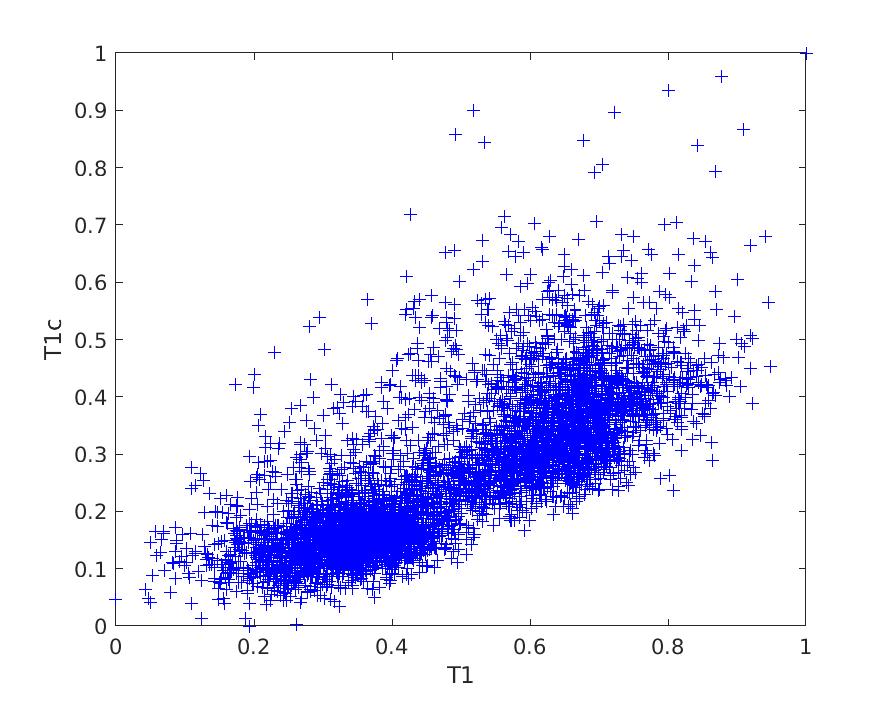}%
}
\hfil
\subfloat[]{\includegraphics[width=1.5in]{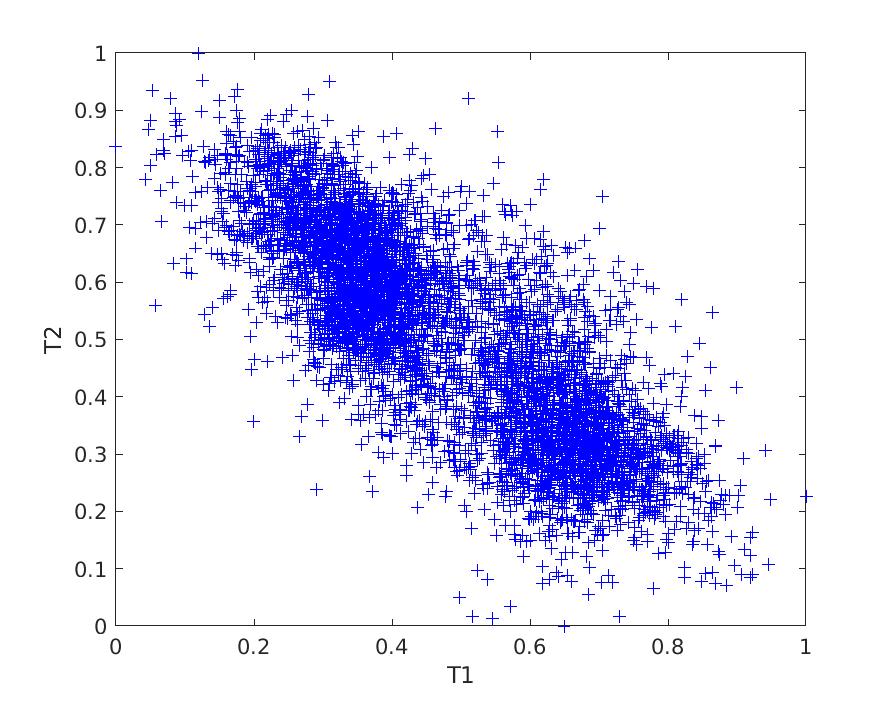}%
}
\hfil
\subfloat[]{\includegraphics[width=1.5in]{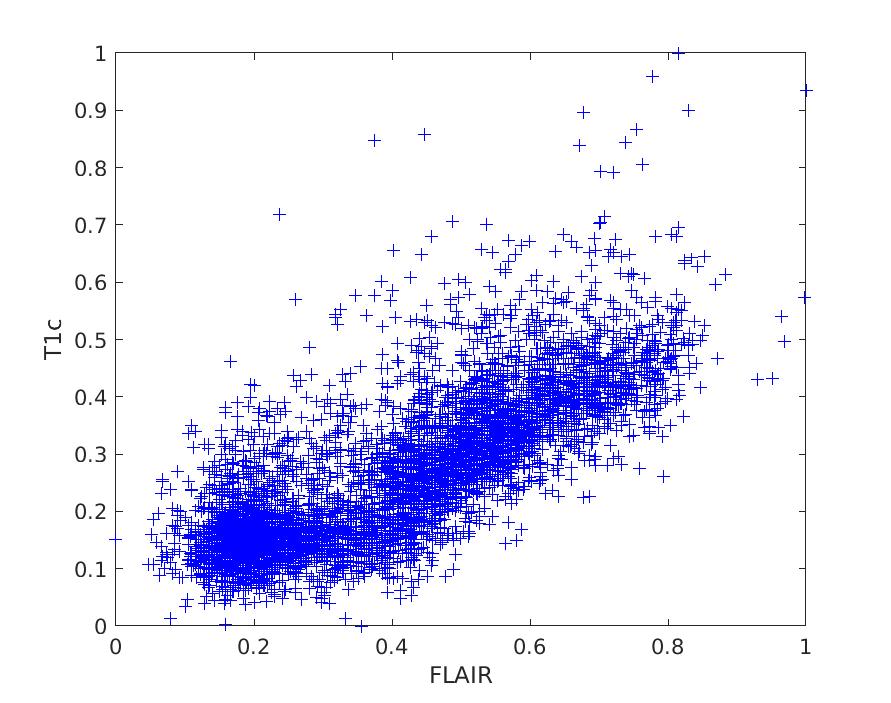}%
}
\hfil
\subfloat[]{\includegraphics[width=1.5in]{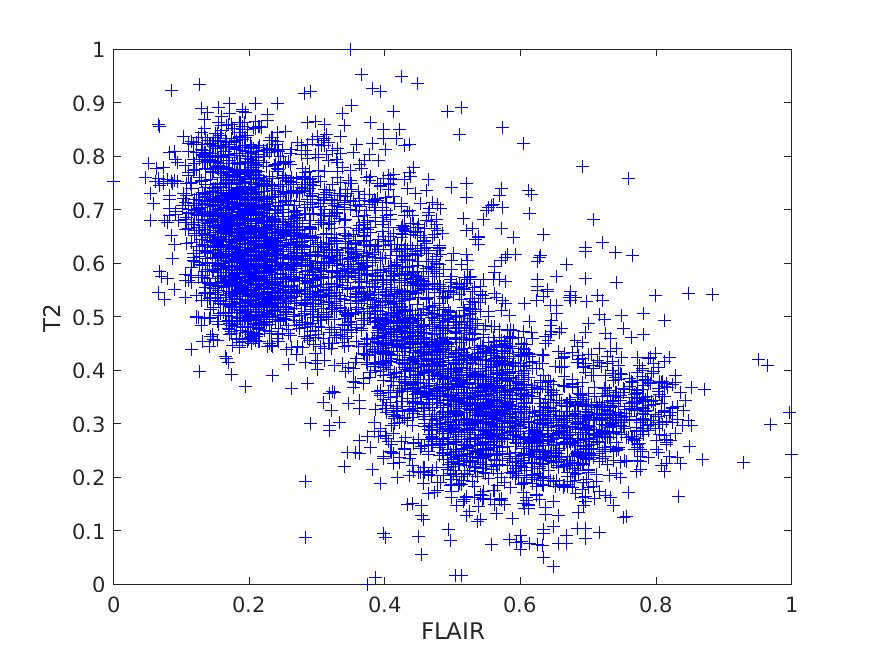}%
}
\hfil
\subfloat[]{\includegraphics[width=1.5in]{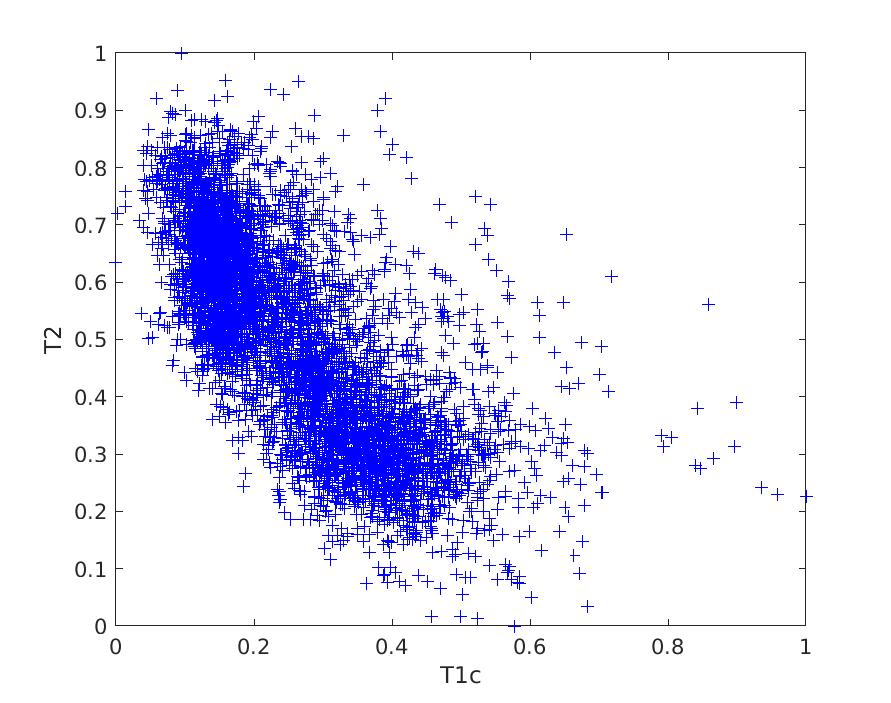}%
}
\caption{Joint intensity distributions of MR images: (a) T1-FLAIR, (b) T1-T1c, (c) T1-T2, (d) FLAIR-T1c, (e) FLAIR-T2, (f) T1c-T2. The intensity of the first modality is read on abscissa axis and that of the second modality on the ordinate axis.}
\label{fig3}
\end{figure}

\begin{figure*}[htb]
\centering
\includegraphics[width=\textwidth]{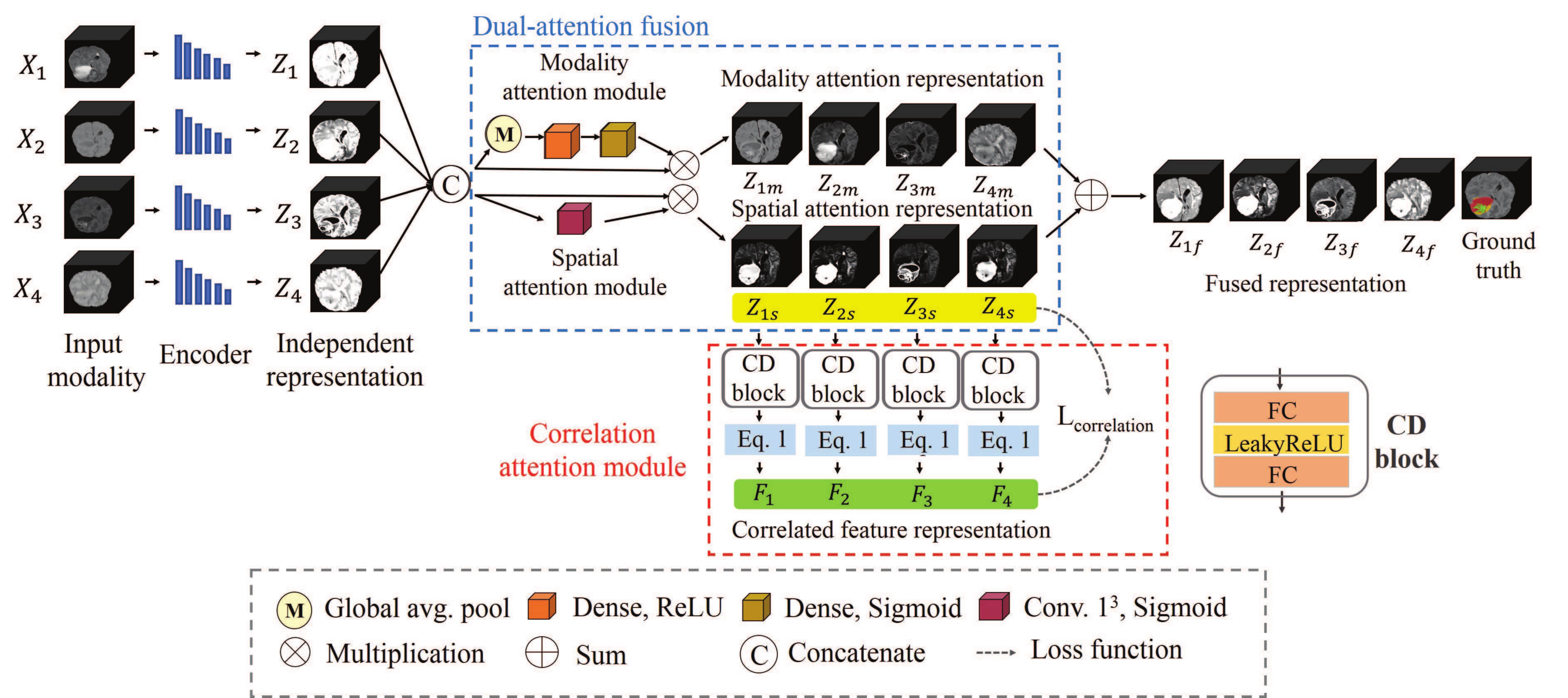}
\caption{Architecture of the tri-attention fusion strategy. The individual feature representations ($Z_1$, $Z_2$, $Z_3$, $Z_4$) are first concatenated, then they are re-weighted by dual-attention fusion block along modality attention module and spatial attention module to achieve the modality attention representation $Z_{im}$ and spatial attention representation $Z_{is}$. In addition, the correlation attention module is used to constrain the spatial-attention representations to learn segmentation-related representation. Finally, the $Z_{im}$ and $Z_{is}$ are added to obtain the fused feature representation $Z_{if}$.}
\label{fig4}
\end{figure*}

Inspired by a fact that, there is strong correlation between multi MR modalities, since the same brain tumor region is observed by different modalities~\cite{lapuyade2017segmenting}. From \autoref{fig3} presenting joint intensities of the MR images, we can observe a strong correlation (not always linear) in intensity distribution between each pair of modalities. To this end, it's reasonable to assume that a strong correlation also exists in latent feature representation between modalities. Therefore, we proposed a correlation attention module and integrated it to the dual-attention fusion block to achieve a tri-attention fusion block. It's used to exploit and utilize the multi-source correlation between modalities, the architeccture is depicated in \autoref{fig4}.

The input modality $\{X_i, ... , X_n\}$, where $n = 4$, is first input to the independent encoders (with learning parameters $\theta$ including the number of the filters and dropout rate) to learn the modality-specific representation $Z_i$. Then, a dual-attention fusion block is used. It takes the concatenation of the independent feature representations as input to produce the modality-weight and spatial-weight, respectively. And the two weights are multiplied with the input feature representation to obtain the modality-attention feature representation $Z_{im}$ and spatial-attention feature representation $Z_{is}$, respectively. Finally, the learned fused feature representation is obtained by adding the modality-attention feature representation and spatial-attention feature representation.

The obtained spatial-attention feature representation $Z_{is}$ is passed to the Correlation Description (CD) block consisting of two fully connected layers and LeakyReLU, it maps the spatial-attention feature representation $Z_{is}$ to a set of independent parameters $\Gamma_i =\{\alpha_i, \beta_i, \gamma_i\}$, $i=1,... ,n$. Finally, the correlated representation of $i$ modality  $F_i$ can be obtained via correlation expression (\autoref{eq1}). 

\begin{equation}
    F_i = \alpha_i \odot Z_{is}^2+ \beta_i \odot Z_{is} +\gamma_i
\label{eq1}
\end{equation}

It is noted that the nonlinear correlation expression we proposed in this work is specific to our work. However, the proposed correlation description block can be generally integrated to any multi-source correlation problem, and the specific correlation expression will depend on the application. In addition, we compare and discuss why the simplest linear correlation expression is not good for this work in Section \ref{6-1}.

Then, the Kullback–Leibler divergence (\autoref{eq2}) is used to measure the divergence between the estimated correlated feature representation of $i$ modality and the spatial-attention feature representation of $j$ modality, which enables the segmentation network to learn the latent correlated features which are more relevant for segmentation. To make it clear, we take T1 modality ($X_1$) and T1c modality ($X_3$) as example, since there exists a correlation between the two modalities, the spatial attention module is first used to obtain the two spatial-attention feature representations of T1 modality ($Z_{1s}$) and T1c modality ($Z_{3s}$), then the correlated feature representation ($F_1$) of modality T1 can be obtained by CD block and \autoref{eq1}. Finally, the KL based correlation loss function is applied to constrain the two distributions ($F_1$ and $Z_{3s}$) to be as close as possible. For our task, it is the divergence between two feature representation distributions that needs to be measured. To this end, we choose a simple and widely used f-divergence function, Kullback–Leibler divergence. It will be interesting to test in future other f-divergence functions, such as Hellinger distance.
\begin{equation}
 L_{correlation} = \sum_{x\in X} P(x)log\frac{P(x)}{Q(x)}
\label{eq2}
\end{equation}

\noindent where $P$ and $Q$ are probability distributions of spatial-attention feature representation ($Z_{js}$) of modality $j$ and correlated feature representation ($F_i$) of modality $i$, ($i\neq j$), respectively, which defined on the same probability space $X$.

From \autoref{fig4}, we can observe the characteristics of the target tumors in the four independent feature representations ($Z_1$, $Z_2$, $Z_3$, $Z_4$) are not obvious. However, the modality attention module can stand out the different attributes of the modalities to provide complementary information. For example, the FLAIR modality ($Z_{2m}$) highlights the whole tumor region and T1c modality ($Z_{3m}$) stands out the tumor core region (red and green). In the spatial attention module, all the positions related to the target tumor regions are highlighted. In this way, we can discover the most relevant characteristics between modalities. Furthermore, the proposed tri-attention fusion strategy can be directly adapted to any multi modal (if existing a correlation relationship) fusion problem. 

\section{Data and Implementation Details}
\label{sec4}
\subsection{Data}
\label{4-1}
The datasets used in the experiments come from BraTS  2018  dataset. The  training  set  includes  285  patients,  each  patient  has four image modalities including T1, T1c, T2 and FLAIR. Following the challenge, four intra-tumor structures have been grouped into three mutually inclusive tumor regions: (a) whole tumor (WT) that consists of all tumor tissues, (b) tumor core (TC) that consists of the enhancing tumor, necrotic and non-enhancing tumor core, and (c) enhancing tumor (ET). The provided data have been pre-processed by organisers: co-registered to the same anatomical template, interpolated to the same resolution ($1mm^3$) and skull-stripped. The ground truth have been manually labeled by experts. We did additional pre-processing with a standard procedure. The N4ITK \cite{avants2009advanced} method is used to correct the distortion of MRI data, and intensity normalization is applied to normalize each  modality of each patient. To exploit the spatial contextual information of the image, we use 3D images, crop and resize them from $155\times240\times240$ to $128\times128\times128$.

\subsection{Implementation Details}
\label{4-2}
Our network is implemented in Keras with a single Nvidia GPU Quadro P5000 (16GB). The models are optimized using the Adam optimizer(initial learning rate  =  5e-4) with a decreasing learning rate factor 0.5 with patience of 10 epochs, to avoid over-fitting, early stopping is used when the validation loss isn't improved for 50 epoch. We randomly split the dataset into 80\% training and 20\% testing.  

\subsection{The choices of loss function}
\label{4-3}

For segmentation, we use dice loss to evaluate the overlap rate of prediction results and ground truth.

\begin{equation}
    \ L_{dice}=1-2\frac{\sum_{i=1}^C\sum_{j=1}^N p_{ij} g_{ij}+\epsilon} {\sum_{i=1}^C\sum_{j=1}^N (p_{ij} + g_{ij})+\epsilon}
\end{equation}

\noindent where $N$ is the set of all examples, $C$ is the set of the classes, $p_{ij}$ is the probability that pixel $i$ is of the tumor class $j$, the same is true for $g_{ij}$, and $\epsilon$ is a small constant to avoid dividing by 0.

The network is trained by the overall loss function as follow:

\begin{equation}
    \ L_{total}= L_{dice}+\lambda \sum_{i=1}^n L_{correlation_n}
\end{equation}

\noindent where $\lambda$ is the trade-off parameter weighting the importance of each component, $n$ denotes the number of correlation pair, in this work, we used three most correlated pairs: T1-T1c, T1-T2, T2-FLAIR.

\begin{figure*}[!t]
\centering
\includegraphics[width=\columnwidth]{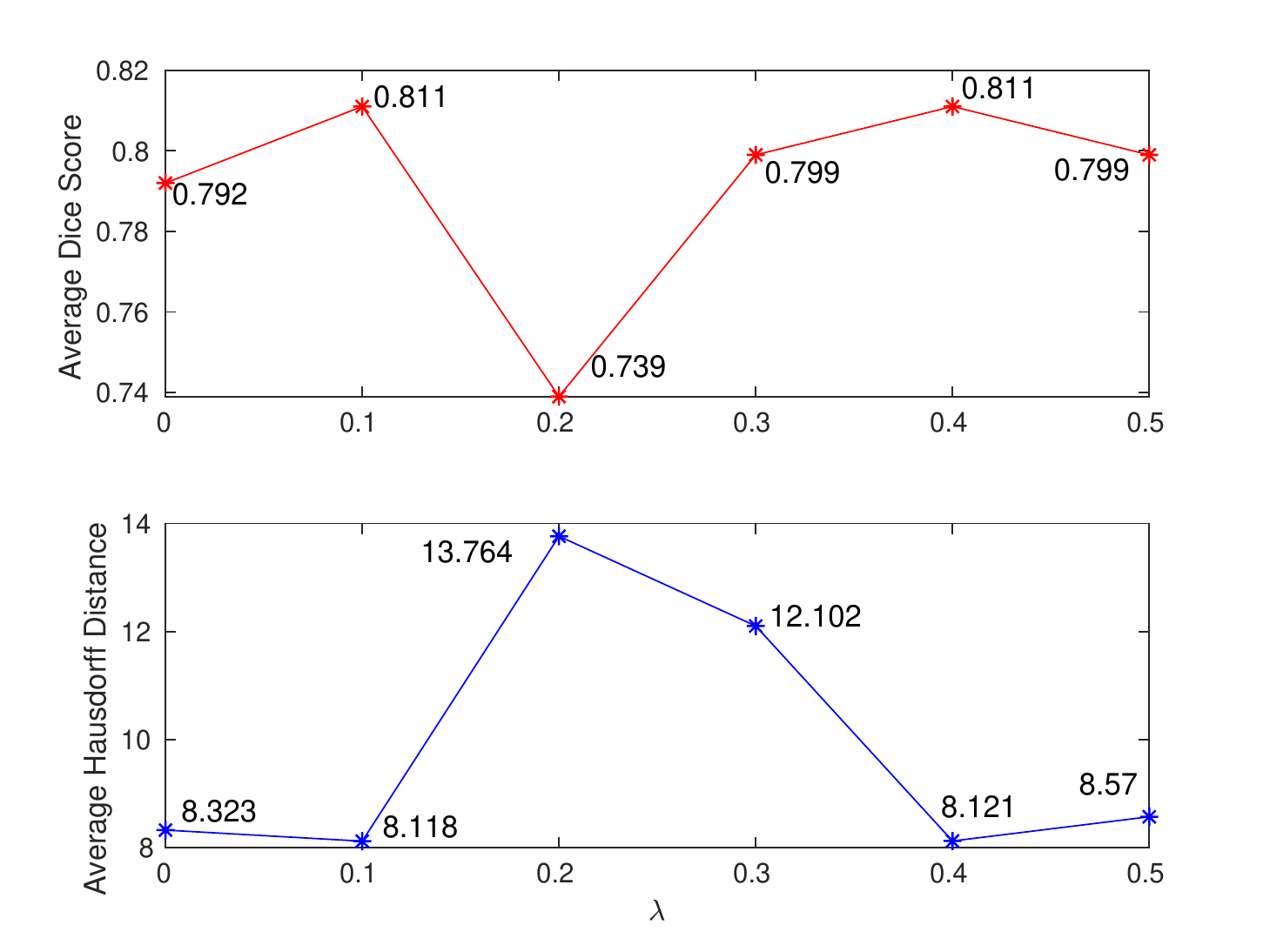}%
\caption{Comparison of different weight coefficients in the loss function. Average Dice Score vs $\lambda$ and Average Hausdorff Distance vs $\lambda$.}
\label{fig5}
\end{figure*}

We did a grid search between $(0,0.5)$ to determine the optimal value for the weight coefficient $\lambda$, \autoref{fig5} shows the comparison of average Dice Score and Hausdorff Distance between different weight coefficients, we found that $\lambda=0.1$ can achieve the best segmentation results.

\subsection{Evaluation metrics}
\label{4-4}
To evaluate the proposed method, two evaluation metrics: Dice Score and Hausdorff distance are used to obtain quantitative measurements of the segmentation accuracy:

\noindent 1) Dice Score: It is designed to evaluate the overlap rate of prediction results and ground truth. It ranges from 0 to 1, and the better predict result will have a larger Dice value.

\begin{equation}
 Dice = \frac {2TP}{2TP+FP+FN}
\end{equation}

\noindent where $TP$ represents the number of true positive voxels, $FP$ represents the number of false positive voxels, and $FN$ represents the number of false negative voxels. 

\noindent 2) Hausdorff Distance (HD): It is computed between boundaries of the prediction results and ground-truth, it is an indicator of the largest segmentation error. The better predict result will have a smaller HD value.

\begin{equation}
  HD=\max\{sup_{r_\in{\partial R}}d_m(s,r),sup_{s_\in\partial S}d_m(r,s)\}
\end{equation}

\noindent where $\partial S$ and $\partial R$ are the sets of tumor border voxels for the predicted and the real annotations, and $d_m(v,v)$ is the minimum of the Euclidean distances between a voxel $v$ and voxels in a set $v$.

\section{Experiment Results}
\label{sec5}
We conduct a series of comparative experiments to demonstrate the effectiveness of our proposed method and compare it to other approaches. In Section \ref{B}, we first perform an ablation experiment to see the importance of our proposed components and demonstrate that adding the proposed components can enhance the segmentation performance. In Section \ref{D}, we compare our method with the state-of-the-art methods. In Section \ref{5-2}, the qualitative experiment results further demonstrate that our proposed method can achieve a promising segmentation result.

\begin{figure}[htb]
\centering
\includegraphics[width=\columnwidth]{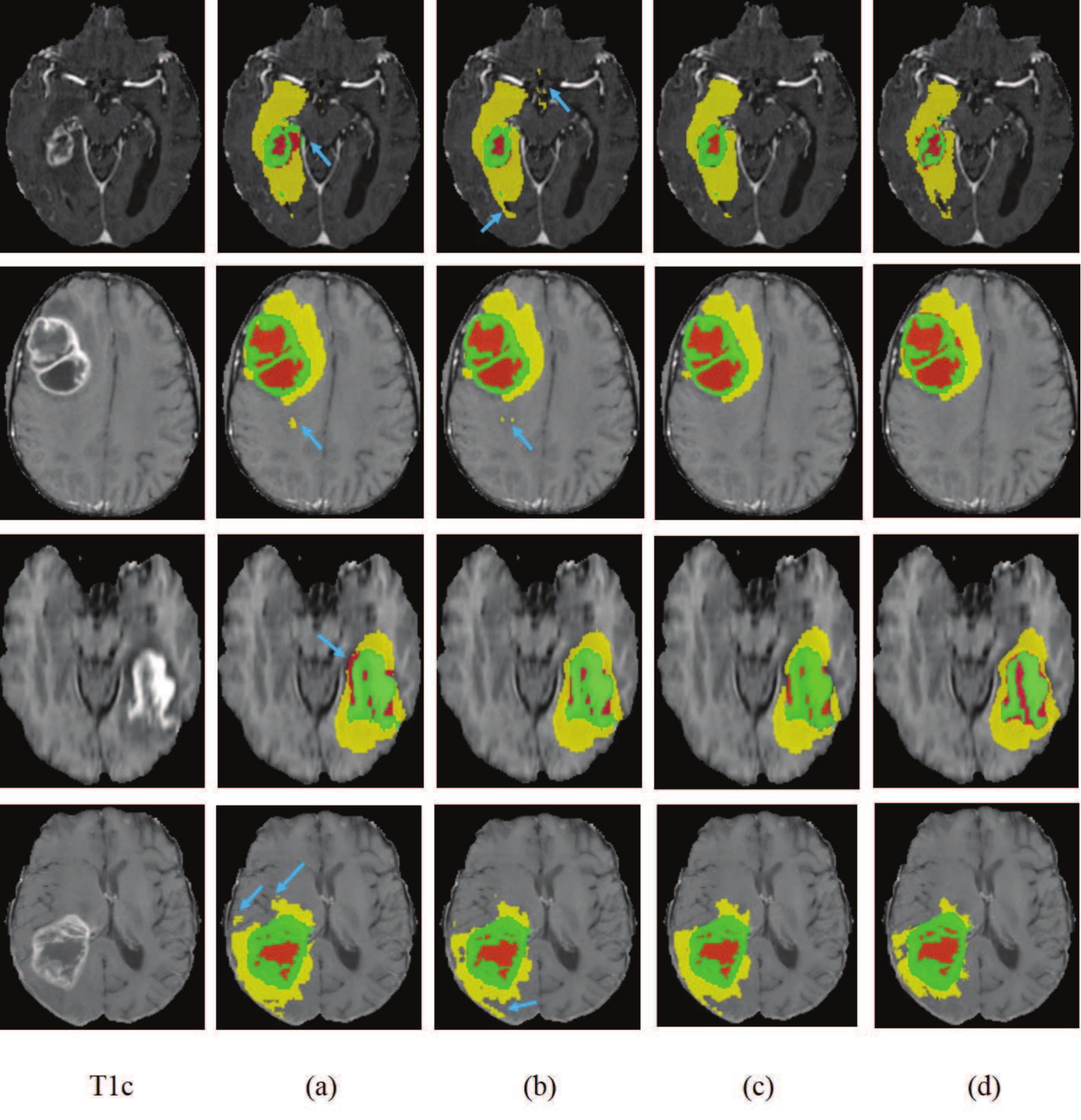}
\caption{Visualization of several segmentation results. (a) Baseline (b) Baseline with dual attention fusion (c) Baseline with tri-attention fusion (d) Ground truth. Red: necrotic and non-enhancing tumor core; Yellow: edema; Green: enhancing tumor. Blue arrow emphasizes the mis-segmentation of the methods.}
\label{fig6}
\end{figure}

\begin{figure}[htb]
\centering
\includegraphics[width=\textwidth]{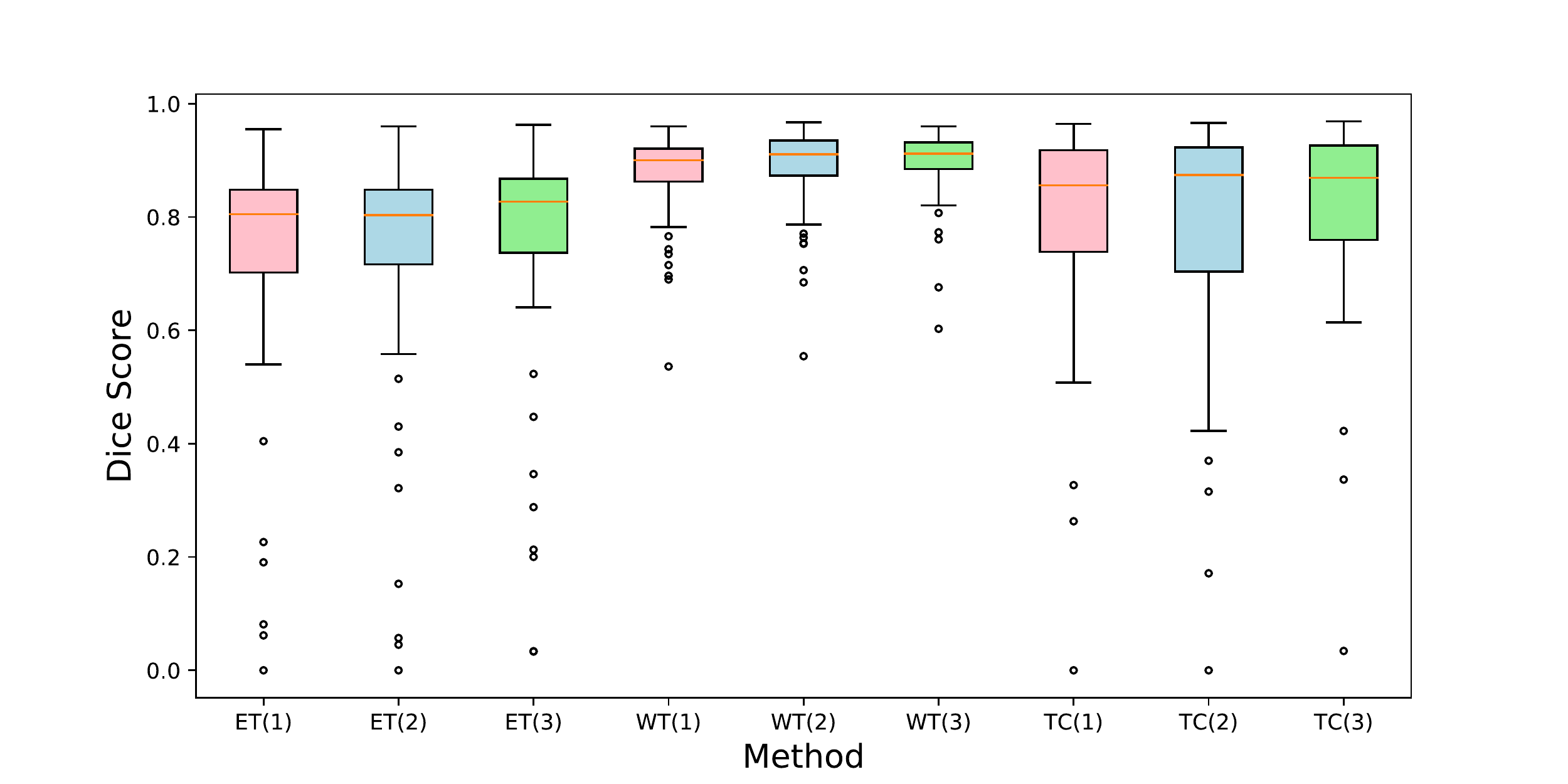}
\caption{Box plots of Dice Score for the three compared methods in \autoref{tab1} with regard to the three tumor regions: enhancing tumor (ET), whole tumor (WT) and tumor core (TC). Method (1) is shown in pink, method (2) in blue and method (3) in green.}
\label{fig7}
\end{figure}

\begin{figure}[htb]
\centering
\includegraphics[width=\textwidth]{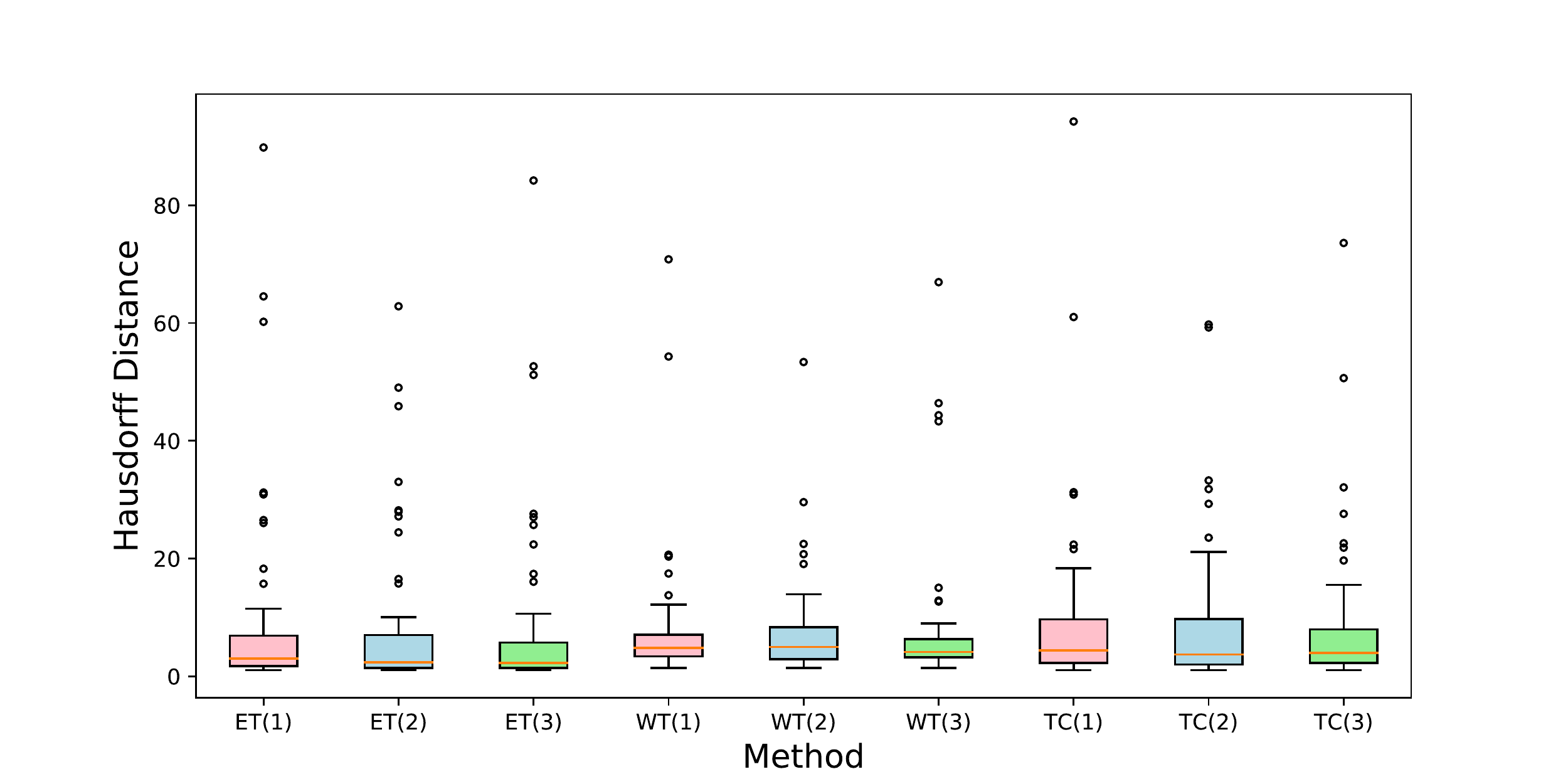}
\caption{Box plots of Hausdorff Distance for the three compared methods in \autoref{tab1} with regard to the three tumor regions: enhancing tumor (ET), whole tumor (WT) and tumor core (TC). Method (1) is shown in pink, method (2) in blue and method (3) in green.}
\label{fig8}
\end{figure}

\subsection{Quantitative analysis}
\label{5-1}
To prove the effectiveness of our network, we first did an ablation experiment to see the effectiveness of our proposed components, and then we compare our method with the state-of-the-art methods. All the results are obtained by online evaluation platform\footnote{https://ipp.cbica.upenn.edu/}.

\subsubsection{Effectiveness of individual modules}
\label{B}
To assess the performance of our method, and see the importance of the proposed components in our network, including dual attention fusion strategy and correlation attention module, we did an ablation experiment, our network without the dual attention fusion and correlation attention module is denoted as baseline. From \autoref{tab1}, we can observe the baseline method achieves Dice Score of 0.726, 0.867, 0.764 for enhancing tumor, whole tumor, tumor core, respectively. When the dual attention fusion strategy is applied to the network, we can see an increase of Dice Score, Hausdorff Distance across all tumor regions with an average improvement of 0.76\% and 6.44\% compared to the baseline, respectively. The major reason is that the proposed fusion block can help to emphasize the most important representations from the different modalities across different positions in order to boost the segmentation result. In addition, another advantage of our method is using the correlation attention module in the lowest layer, which can constrain the encoders to discover the latent multi-source correlation representation between modalities and then guide the network to learn correlated representation to achieve a better segmentation. From the results, we can observe that with the assistance of correlation attention module, the network can achieve the best Dice Score of 0.75, 0.887 and 0.796, Hausdorff Distance of 7.687, 8.306 for enhancing tumor, whole tumor, tumor core, respectively with an average improvement of 3.18\% and 8.75\% relating to the baseline. In addition, we visualized the box plots of the Dice Score and Hausdorff Distance for the three compared methods in \autoref{fig7} and \autoref{fig8}. It can be observed that our proposed method not
only has a higher accuracy but also a smaller standard deviation than the other two compared methods in the terms of Dice Score and Hausdorff Distance. The results in \autoref{tab1}, \autoref{fig7} and \autoref{fig8} demonstrate the effectiveness of each proposed component and our proposed network architecture can perform well on brain tumor segmentation.

\begin{table}[]  
\centering
\caption{Evaluation of our proposed method on Brats 2018 training dataset, (1) Baseline (2) Baseline + Dual attention fusion (3) Baseline + Tri-attention fusion, ET, WT, TC denote enhancing tumor, whole tumor and tumor core, respectively. Avg denotes the average results on the three tumor regions, bold results denote the best results.}
\label{tab1}
\vspace{0.3cm}
\renewcommand{\arraystretch}{1.3}%
\resizebox{\textwidth}{!}{%
\begin{tabular}{ccccccccc}
\hline
{Methods}&\multicolumn{4}{c}{Dice Score} & \multicolumn{4}{c}{Hausdorff Distance (mm)} \\  \cline{1-9}
& ET & WT & TC &Avg & ET & WT & TC &Avg\\ \hline
(1) & 0.726 & 0.867 & 0.764 & 0.786&8.743 & 8.463 & 9.482 &8.896\\
(2) & 0.733 & 0.879 & 0.765 &0.792& 8.003 &\textbf{7.813} & 9.153&8.323 \\
(3) & \textbf{0.750} & \textbf{0.887} & \textbf{0.796}  &  \textbf{0.811} & \textbf{7.687}& 8.306 & \textbf{8.362} & \textbf{8.118}\\
\hline
\end{tabular}}
\end{table}

\subsubsection{Comparisons with the state-of-the-art}
\label{D}
Since access to the testing set of BraTS 2018 was closed after the challenge, we compare our proposed method with the state-of-the-art methods on BraTS 2018 online validation set, which contains 66 images of patients without the ground-truth. We first predict the segmentation results on local machine and then submitted on the online evaluation platform to obtain the evaluation results. \autoref{tab2} shows the comparison results. The experiment results of methods \cite{zhao2018deep} and \cite{kamnitsas2017efficient} are cited from \cite{akil2020fully}. We also did a computational complexity comparison between these state-of-the-art methods, including the data dimension, input size, number of network layers, number of initial convolution filter, data augmentation, post-processing, used GPU and training time, shown in \autoref{tab3}.

(1) Zhao et al. \cite{zhao2018deep} proposed to integrate Fully Convolutional Neural Networks (FCNNs) and Conditional Random Fields (CRFs) in a unified framework, where three segmentation models using 2D image patches and slices are trained in axial, coronal and sagittal views, respectively, and they are combined to segment brain tumors using a voting based fusion strategy.

(2) Kamnitsas et al. \cite{kamnitsas2017efficient} introduced a dual pathway 3D convolutional neural network to incorporate both local and larger contextual information for brain tumor segmentation. In addition, they used a 3D fully connected CRF as the post-processing to remove the false positives.

% introduced EMMA, an ensemble of widely varying CNNs. By combining a heterogeneous collection of networks, the proposed model is insensitive to independent failures of CNN components and thus generalises well, which won the first place in BraTS 2017 challenge.

(3) Hu et al. \cite{hu2018brain} proposed the multi-level up-sampling network (MU-Net) for automated segmentation of brain tumors, where a novel global attention (GA) module is used to combine the low level feature maps obtained by the encoder and high level feature maps obtained by the decoder.

(4) Gates et al. \cite{gates2018glioma} applied a multi-scale convolutional neural network based on the DeepMedic \cite{kamnitsas2017efficient} to segment brain tumor.

(5) Tuan et al. \cite{tuan2018brain} proposed using Bit-plane to generate a series of binary images by determining significant bits. Then, the first U-Net used the significant bits to segment the tumor boundary, and the other U-Net utilized the original images and images with least significant bits to predict the label of all pixel inside the boundary. 

(6) Hu et al. \cite{hu2018hierarchical} introduced the 3D-residual-Unet architecture. The network comprises a context aggregation pathway and a localization pathway, which encoder abstract representation of the input, and then recombines these representations with shallower features to precisely localize the interest domain via a localization path.

(7) Myronenko et al. \cite{myronenko20183d} proposed a 3D MRI brain tumor segmentation using autoencoder regularization, where a variational autoencoder branch is added to reconstruct the input image itself in order to regularize the shared decoder and impose additional constraints on its layers. 

From \autoref{tab2}, we first observe that the U-Net based network \cite{hu2018brain, tuan2018brain, hu2018hierarchical, myronenko20183d} can achieve better results than the CNN based network \cite{zhao2018deep, kamnitsas2017efficient, gates2018glioma}. We explain that the skip connections in the U-Net can combine the high-level semantic feature maps from the decoder and corresponding low-level detailed feature maps from the encoder, which allows the network to learn more useful feature information to improve the segmentation. In addition, the best result in BraTS 2018 Challenge is from \cite{myronenko20183d}, which achieves 0.814, 0.904 and 0.859 in terms of Dice Score on enhancing tumor, whole tumor and tumor core regions, respectively. However, from \autoref{tab3}, we can observe that it uses 32 initial convolution filters and a lot of memories (NVIDIA Tesla V100 32GB GPU is required) to train the model, which is computationally expensive. While our method used only 8 initial filters, a 16GB GPU is sufficient to conduct our experiments, and our network uses less training time. And from \autoref{tab2}, it can be observed that our proposed method can yield a competitive results in terms of Dice Score and Hausdorff Distance across all the tumor regions. The main advantage of our method is that it takes into account the multi-source correlation in brain MRI to find those features which are relevant to obtain good segmentation. The proposed correlation attention module is a general one which can be applied to other multimodal fusion applications if a correlation exists between them. Furthermore, compared with other methods, \cite{hu2018hierarchical} has better Dice Score and Hausdorff Distance on enhancing tumor, while our method uses smaller input size but one more layer, and finally achieves a better average Dice Score on all the tumor regions with an improvement of 3.84\%, and it can also obtain an average improvement of 7.5\% for Hausdorff Distance. 

% \textcolor{red}{We also did a computational complexity comparison between these state-of-the-art methods, including the data dimension, input size, number of network layers, number of initial filter, data augmentation, post-processing, used GPU and training time shown in \autoref{tab3}. We can observe that the larger input size and more layers can attribute to the better segmentation results. Comparing with \cite{hu2018hierarchical}, our proposed method used smaller input size but one more layer, and finally achieved better results in the terms of average Dice Score and Hausdorff Distance. When compared with the best method \cite{myronenko20183d}, which used 32 initial convolution filters. Our method used 8 initial convolution filters, which requires less GPU, and our network used less training time, it can also achieve promising results.} 

% With the same GPU equipment and the same segmentation performance in the terms of average Dice Score, \cite{zhao2018deep} used more training time than \cite{kamnitsas2017efficient}. It is because \cite{zhao2018deep} used more complex network architecture, which integrated CRF into the FCNN. While \cite{kamnitsas2017efficient} used CRF as the post-processing. From \cite{hu2018brain} and \cite{tuan2018brain}, 
\begin{table}[]
\centering
\caption{Comparison of different methods on BraTS 2018 validation dataset, ET, WT, TC denote  enhancing tumor, whole tumor, tumor core, respectively. Avg denotes the average results on the three tumor regions, bold results denote the best results, underline results denote the second best results. ”-” indicates that the information is not provided in the published paper.}
\label{tab2}
\vspace{0.3cm}
\renewcommand{\arraystretch}{1.3}%
\resizebox{\textwidth}{!}{%
\begin{tabular}{ccccccccc}
\hline
{Methods} & \multicolumn{4}{c}{Dice Score} & \multicolumn{4}{c}{Hausdorff Distance (mm)} \\ \cline{1-9}
 & ET & WT & TC & Avg& ET & WT & TC & Avg\\ \hline
% \cite{ellwaa2016brain}& 0.56 & 0.82 & 0.72 &0.7 & - & - & - &-\\
\cite{zhao2018deep}& 0.62 & 0.84 & 0.73 &0.715 & - & - & - &-\\ 
\cite{kamnitsas2017efficient}& 0.629 & 0.847 & 0.67 &0.715& - & - & - &-\\
% \cite{hu2018brain} & 0.66 & 0.87 & 0.72 & 0.75 & 7.56 & 6.73 & 15.74 & 10.01 \\
\cite{hu2018brain} & 0.69 & 0.88 & 0.74 & 0.77 & 6.69 & 4.76 & 10.67 & \underline{7.373} \\
\cite{gates2018glioma} &0.678&0.805&0.685&0.723&14.522&14.415&20.017&16.318\\
\cite{tuan2018brain} & 0.682 & 0.818 & 0.699 & 0.733 & 7.016 & 9.421 & 12.462 & 9.633 \\
\cite{hu2018hierarchical}& \underline{0.719} & 0.856 & 0.769 & 0.781 & \underline{5.5} & 10.843 & \underline{9.985} & 8.776 \\
\cite{myronenko20183d}& \textbf{0.814} & \textbf{0.904} & \textbf{0.859} & \textbf{0.859} & \textbf{3.804} & \textbf{4.483} & \textbf{8.278} & \textbf{5.521} \\
% Proposed  & 0.705 & \underline{0.883} & \underline{ 0.783} &  \underline{0.79} & 7.27 & \underline{5.111} & 10.047 &  \underline{7.476} \\
Proposed  & 0.688 & \underline{0.876} & \underline{ 0.784} &  \underline{0.783} & 6.900 & \underline{6.551} & 10.199 & 7.883\\
\hline
\end{tabular}%
}
\end{table}

\begin{table}[]
\centering
\caption{Computational complexity comparison of different methods. ”-” indicates that the information is not provided in the published paper.}
\label{tab3}
\vspace{0.3cm}
\renewcommand{\arraystretch}{1.3}%
\resizebox{\textwidth}{!}{%
\begin{tabular}{ccccccccc}
\hline
Methods & Data dimension & Input size & Num. layer & Num. initial filter& Data augmentation & Post-processing & GPU & Training Time\\ \hline
\cite{zhao2018deep} & 2D & $65 \times 65$, $33 \times 33$ & 10 & 48& No & Yes & 2G & 288h \\

\cite{kamnitsas2017efficient} & 3D & $25 \times 25 \times 25$, $19 \times 19 \times 19$ & 11 & 30& Yes & Yes & 2G & 72h \\

\cite{hu2018brain} & 2D & $224 \times 224$ & 6 & 64&No & No & - & - \\

\cite{gates2018glioma} & 3D &  $25 \times 25 \times 25$, $19 \times 19 \times 19$  & 11 & 30&No & No & 6G & 96h \\

\cite{tuan2018brain} & 2D & $176 \times 176$ & 5 &64& No & No & - & - \\

\cite{hu2018hierarchical} & 3D & $144 \times 144 \times 144$ & 5 &16& No & No & - & - \\

\cite{myronenko20183d} & 3D & $160 \times 192 \times 128$ & 4 & 32 & Yes & No & 32G & 48h \\

Proposed & 3D & $128 \times 128 \times 128$ & 6 & 8& No & No & 16G & 40h \\ \hline
\end{tabular}%
}
\end{table}

\subsection{Qualitative analysis}
\label{5-2}
In order to evaluate the robustness of our model, we randomly select several examples on BraTS 2018 dataset and visualize the segmentation results in \autoref{fig6}. From \autoref{fig6}, we can observe that the segmentation results are gradually improved when the proposed strategies are integrated, these comparisons indicate that the effectiveness of the proposed strategies. In addition, with all the proposed strategies, our proposed method can achieve the best results.

\section{Discussion}
\label{sec6}
We discuss our method from the following aspects to further demonstrate the effectiveness of our method. Initially, we explore and compare the different correlation expressions in the correlation description block to determine which functional form provides the best fit in Section \ref{6-1}. Subsequently, we analyzed the performance of correlation attention module in different layer of network in Section \ref{6-2}. Finally, we visualize the feature maps of different approaches in Section \ref{6-3} to demonstrate the proposed fusion strategy can improve the segmentation.

\subsection{Performance analysis on correlation expression}
\label{6-1}

\begin{figure}[htb]
\centering
\includegraphics[width=\textwidth]{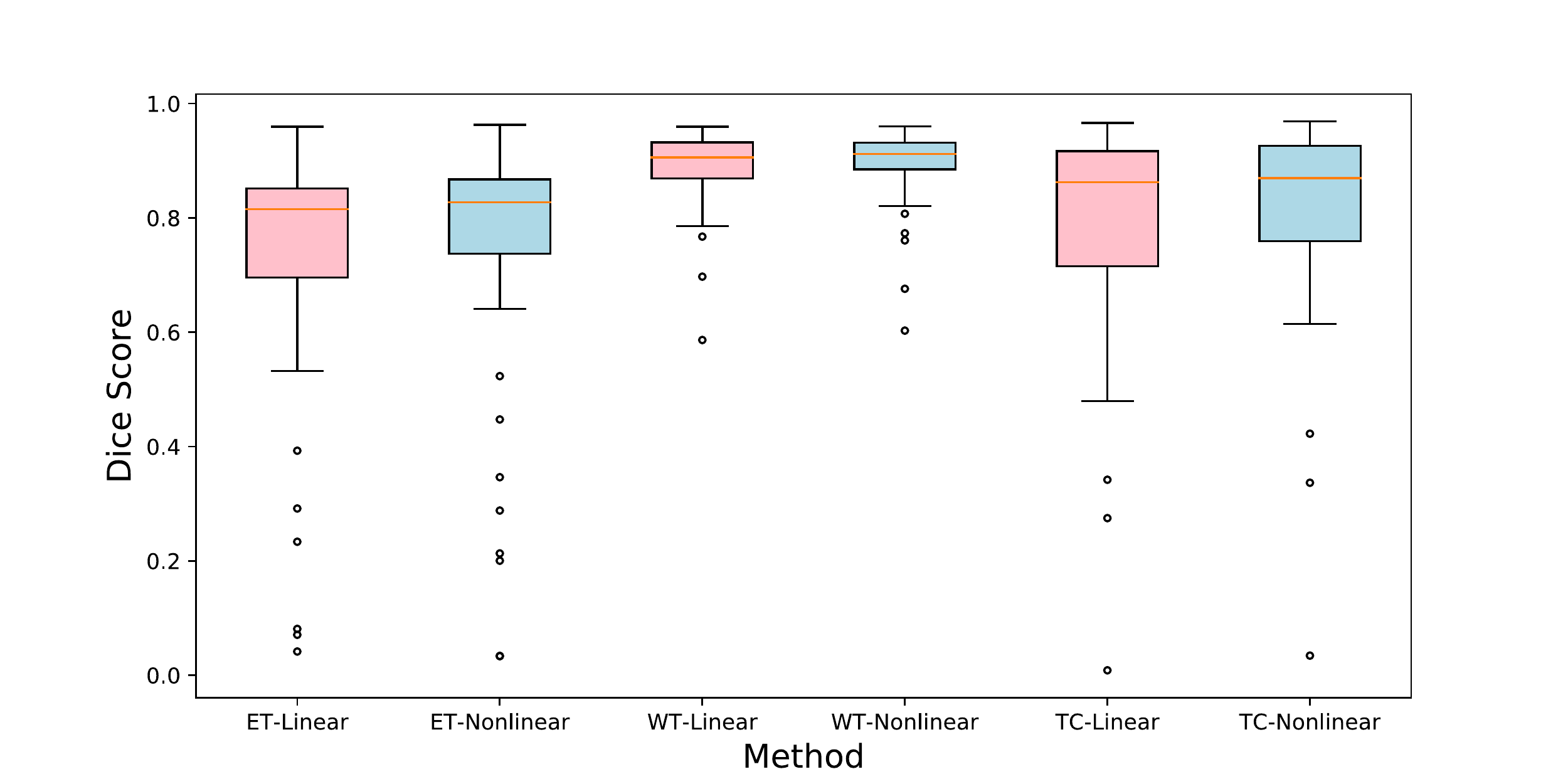}
\caption{Box plots of Dice Score for the two compared correlation expressions in \autoref{tab4} with regard to the three tumor regions: enhancing tumor (ET), whole tumor (WT) and tumor core (TC). Linear expression is shown in pink, Non-linear expression in blue.}
\label{fig9}
\end{figure}

\begin{figure}[htb]
\centering
\includegraphics[width=\textwidth]{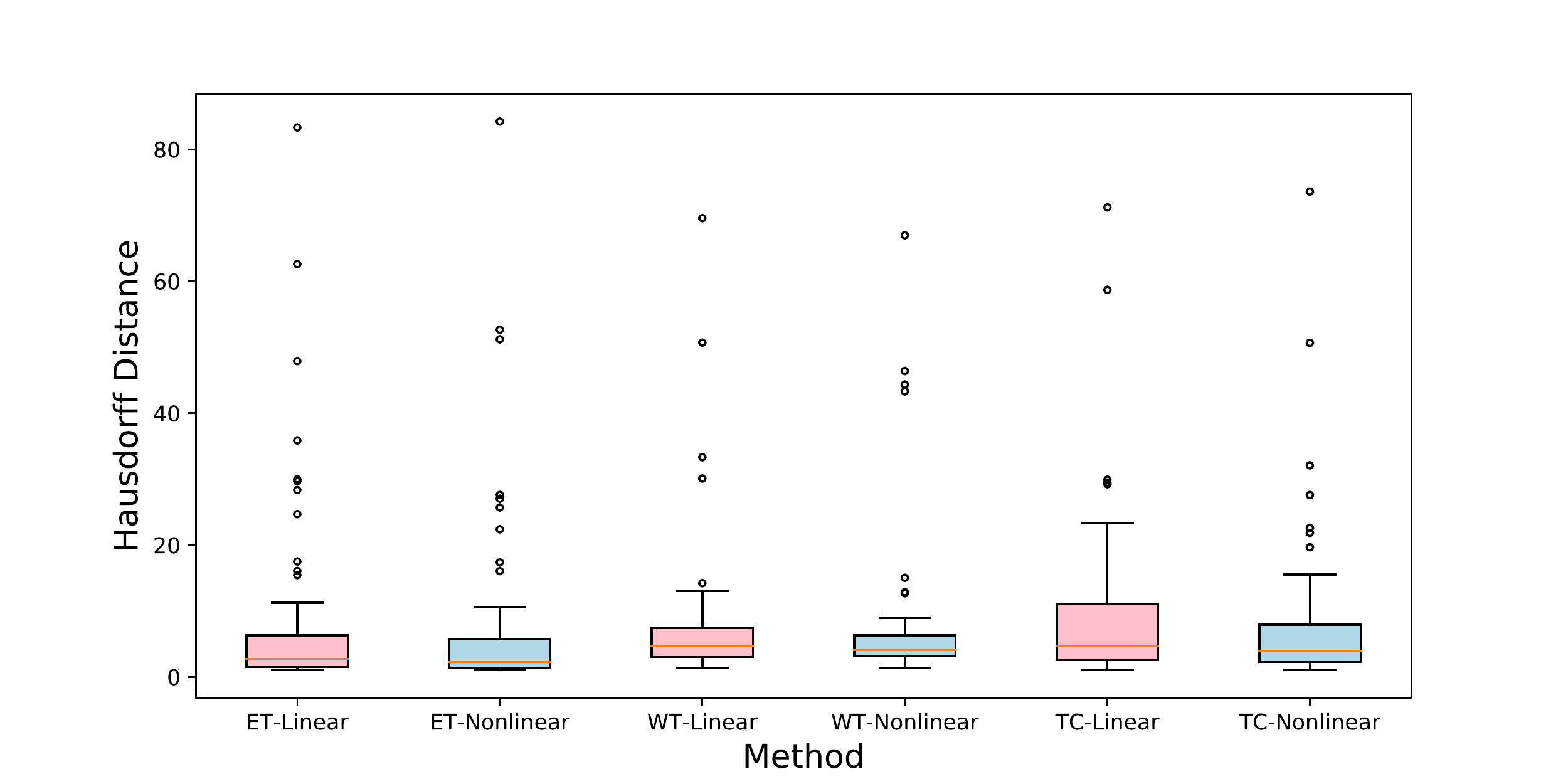}
\caption{Box plots of Hausdorff Distance for the two compared correlation expressions in \autoref{tab4} with regard to the three tumor regions: enhancing tumor (ET), whole tumor (WT) and tumor core (TC). Linear expression is shown in pink, Non-linear expression in blue.}
\label{fig10}
\end{figure}

\autoref{tab4} compares the performance between linear (\autoref{eq3}) and nonlinear (\autoref{eq1}) correlation expression for segmenting brain tumors.  As we can see, the nonlinear correlation expression exhibits clear advantages over the linear correlation expression across all the tumor regions. We explained that the capability is attributed to the complex nonlinear expression, which uses more parameters to fit a feature distribution, giving a better description for the feature distributions so as to guide the network to learn more correlated feature representations for segmentation. In addition, we visualized the box plots of the Dice Score and Hausdorff Distance for the two compared expressions in \autoref{fig9} and \autoref{fig10}. From the two box plots, we can obtain the consistent conclusion that the nonlinear correlation expression can achieve not only a higher accuracy but also a smaller standard deviation than the linear one in the terms of Dice Score and Hausdorff Distance.

\begin{equation}
    F_i = \alpha_i \odot Z_{is} +\gamma_i
\label{eq3}
\end{equation}

\begin{table}[]
\centering
\caption{Comparison of segmentation accuracy of different correlation expression. Avg denotes the average results on the three tumor regions, bold results denote the best results.}
\label{tab4}
\vspace{0.3cm}
\renewcommand{\arraystretch}{1.3}%
\resizebox{\textwidth}{!}{%
\begin{tabular}{ccccccccc}
\hline
{Methods} & \multicolumn{4}{c}{Dice Score} & \multicolumn{4}{c}{Hausdorff Distance (mm)} \\ \cline{1-9}
 & ET & WT & TC & Avg & ET & WT & TC & Avg \\ \hline
Linear & 0.736 & 0.883 & 0.767 & 0.795 &  8.827 & 8.485 &  9.354 &8.889\\
Nonlinear &  \textbf{0.750} &  \textbf{0.887} &  \textbf{0.796} &  \textbf{0.811}& \textbf{7.687}& \textbf{8.306} & \textbf{8.362} & \textbf{8.118} \\ 
\hline
\end{tabular}%
}
\end{table}

\subsection{Performance analysis on correlation attention module}
\label{6-2}
While experimenting with the network architectures, we have tested the addition of the correlation attention module in different layer of network. \autoref{tab5} shows the comparison results, (0) is our method without correlation attention module, which is used as a comparison baseline. As we can see, while setting the correlation attention module in the 4th and 6th layer can achieve a better segmentation results. Then we experimented to set the correlation attention module in both 4th and 6th layers ((7)), while the results aren't improved, therefore, we choose to put it in the 6th layer. Then we tried to put the correlation attention module in more layers, while the correlation attention module in multi-shallower layers ((8)-(12)) did not further improve the segmentation performance. We explained that since each layer represents different abstract feature representation of the input, where deeper levels provide more complex and abstract features, the correlation attention module can guide the most abstract feature distribution to satisfy the correlation relationship in order to improve the segmentation results.

\begin{table*}[]
\centering
\caption{Comparison of segmentation accuracy of correlation attention module in different layer of the network. ET, WT, TC denote
enhancing tumor, whole tumor and tumor core, respectively. Avg denotes the average results on the three tumor regions, bold results denote the best
results.}
\label{tab5}
\vspace{0.3cm}
\renewcommand{\arraystretch}{1.3}%
\resizebox{\textwidth}{!}{%
\begin{tabular}{ccccccccccccccccc}
\hline
 & \multicolumn{6}{c}{Methods} & \multicolumn{4}{c}{Dice Score} & \multicolumn{4}{c}{Hausdorff Distance (mm)}\\ \hline
  & 1st & 2nd & 3rd & 4th & 5th & 6th & ET & WT & TC & Avg & ET & WT & TC & Avg &  \\ \hline
(0) & $-$ & $-$ & $-$ & $-$ & $-$ & $-$  & 0.733& 0.879 & 0.765 & 0.792 & 8.003 & 7.813 &9.153  & 8.323  \\  

(1)& $\surd$ & $\times$ & $\times$ & $\times$ & $\times$ & $\times$ &0.733&0.868&0.744&0.782&8.65&7.603&9.641&8.631 \\

(2)& $\times$ & $\surd$ & $\times$ & $\times$ & $\times$ & $\times$ &0.74&0.878&0.761&0.793&7.978&8.168&9.404&8.517 \\

(3)& $\times$ & $\times$ & $\surd$ & $\times$ & $\times$ & $\times$ &0.741&0.877&0.772&0.797&6.43&\textbf{6.994}&8.119&7.181 \\

(4)& $\times$ & $\times$ & $\times$ & $\surd$ & $\times$ & $\times$ &\textbf{0.762}&0.886&0.776&0.808&\textbf{5.906}&7.516&\textbf{7.809}&\textbf{7.077}   \\

(5)& $\times$ & $\times$ & $\times$ & $\times$ & $\surd$ & $\times$ &0.739&\textbf{0.889}&0.767&0.798&8.071&8.266&10.181&8.839   \\

(6)& $\times$ & $\times$ & $\times$ & $\times$ & $\times$ & $\surd$ &0.75 &0.887 & \textbf{0.796} & \textbf{0.811}  & 7.687 & 8.306 & 8.362 &8.118   \\

(7)& $\times$ & $\times$ & $\times$ & $\surd$ & $\times$ & $\surd$ &0.754&0.886&0.778&0.806&7.677&8.206&9.18&8.354 \\

(8)& $\times$ & $\times$ & $\times$ & $\times$ & $\surd$ & $\surd$ & 0.754 & 0.887 & 0.785 & 0.809 & 7.674 & 7.643 & 8.696 & 8.004 \\

(9)& $\times$ & $\times$ & $\times$ & $\surd$ & $\surd$ & $\surd$  & 0.682 & 0.843 & 0.725& 0.75  &10.282  & 10.161 &11.271  & 10.571  \\

(10)& $\times$ & $\times$ & $\surd$ & $\surd$ & $\surd$ & $\surd$ & 0.695 & 0.822 & 0.699 & 0.739  &10.713  &15.685  & 12.189 & 12.863  \\

(11)& $\times$ & $\surd$ & $\surd$ & $\surd$ & $\surd$ & $\surd$ &0.702 & 0.848 & 0.713 & 0.754 &9.516  &9.449  & 10.64 &9.868  \\

(12) & $\surd$ & $\surd$ & $\surd$ & $\surd$ & $\surd$ & $\surd$  & 0.536 & 0.724 & 0.406 & 0.555 & 17.102 & 30.667 &21.359  & 23.043 \\
\hline
\end{tabular}}
\end{table*}

\subsection{Feature maps visualization}
\label{6-3}

\begin{figure}[htb]
\centering
\includegraphics[width=\columnwidth]{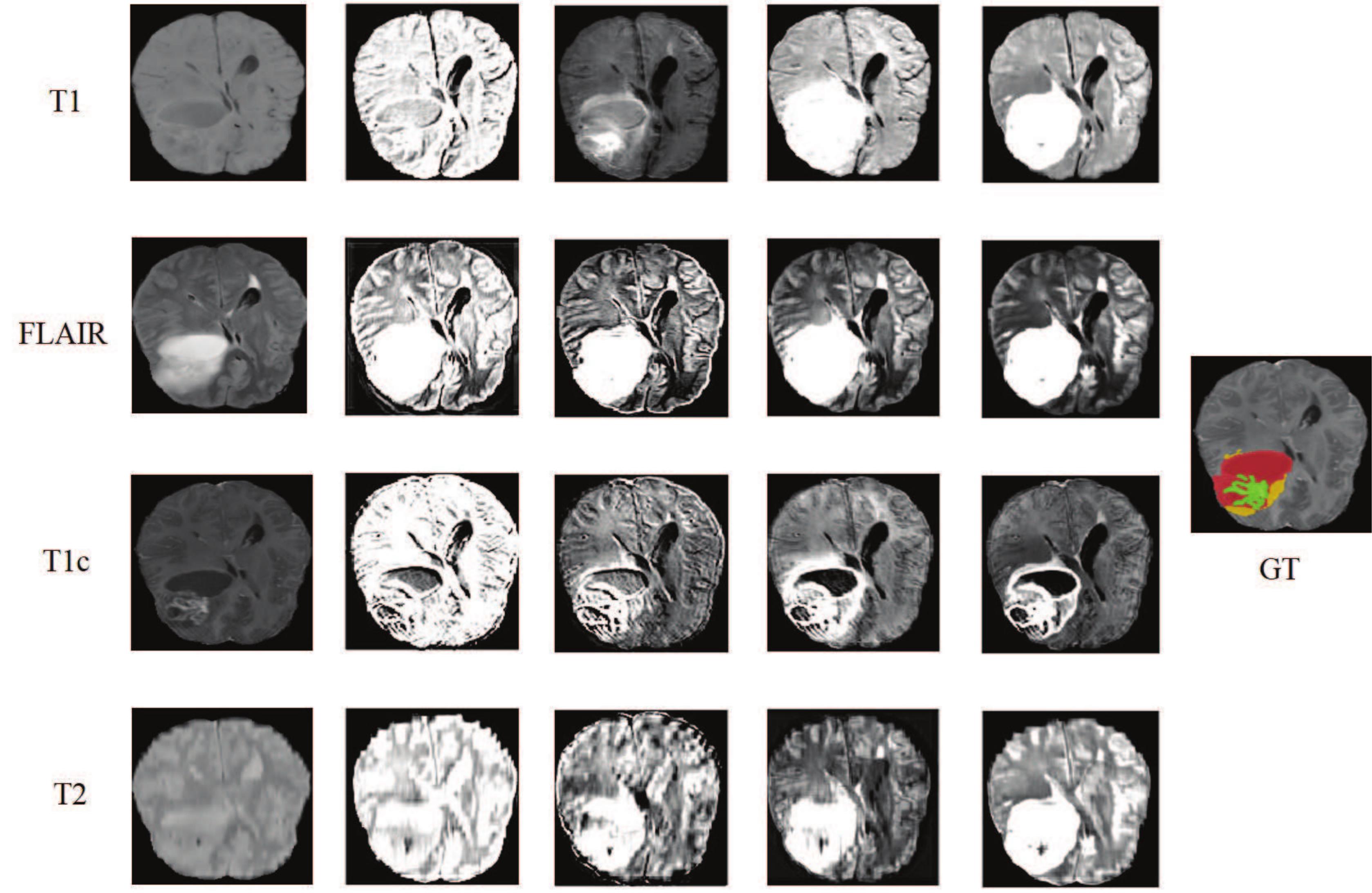}
\caption{Visualization of effectiveness of proposed correlation attention module. First column: input image, second column: baseline, third column: baseline + dual attention module, fourth column: baseline + tri-attention module (added on fused feature representation), fifth column: (added on spatial-attention feature representation), sixth column: ground-truth.}
\label{fig11}
\end{figure}

In this section, we illustrate the advantage of our proposed correlation attention module by visualizing the feature representation maps. We select an example to show the feature representation maps of the first layer from four modalities in \autoref{fig11}. We denote our proposed method without any fusion strategy as baseline, the first column: input modality, the second column: baseline, third column: 'baseline + dual attention module', the fourth column: 'baseline + tri-attention module (added on the fused feature)', the fifth column: 'baseline + tri-attention module (added on the spatial attention feature)', and the sixth column: ground truth. From \autoref{fig11}, we can observe that compared to the baseline, the attention mechanism (column: 3rd, 4th, 5th) allows to highlight feature representations related to brain tumor regions, especially when correlation is taken into account (column: 4th and 5th). In fact, the correlation attention module helps to enhance the fused modality-spatial feature representation for images with fewer information in the tumor region, such as T1 and T2 images. 

To further investigate the contribution of the correlation attention module, we used it to guide the fused feature representations (column: 4th) and spatial-attention feature representations (column: 5th), respectively. From \autoref{fig11}, we can observe that the correlation attention module added on the spatial-attention feature representations (column: 5th) can further stand out the interested tumor regions for segmentation, and the fuzzy contour becomes clear. We explained that the spatial attention module can help the network to extract the feature representations relating the tumor positions. In conclusion, the correlation attention module can constrain the network to emphasize the interested tumor region for segmentation, revealing that the addition of correlation attention module in the network encourages better segmentation results.

\section{Conclusion}
\label{sec7}
In this paper, we proposed a tri-attention fusion guided 3D multi-modal brain tumor segmentation network, where the architecture demonstrated their segmentation performances in multi-modal MR images of glioma patients. 

To take advantage of the complimentary information from different modalities, the multi-encoder based network is used to learn modality-specific feature representation. Considering the correlation between MR modalities can help the segmentation, a tri-attention fusion block is proposed, which consists of a modality attention module, a spatial attention module and a correlation attention module. The modality attention module is used to distinguish the contribution of each modality, and the spatial attention module is used to extract more useful spatial information to boost the segmentation result. Since there is a strong correlation between modalities, a correlation description block is used to represent the multi-modal correlation, a correlation based constraint is introduced to the correlation attention module to guide the network to learn the most correlated feature representation to improve the segmentation. In conclusion, the proposed tri-attention fusion strategy utilized the complimentary information between modalities to encourage the network to learn more useful feature representation to boost the segmentation result.

The advantages of our proposed network architecture (1) The experiment results evaluated on the two metrics (Dice Score and Hausdorff Distance) demonstrate that our proposed method gives an accurate result for the segmentation of brain tumors and its sub-regions even small regions. (2) The architecture are an end-to-end deep leaning approach and fully automatic without any user interventions. (3) The proposed correlation attention module can help the segmentation network to learn correlated feature representations to achieve very competitive results. (4) The proposed correlation attention module can be generalized to other multi-source image processing problem if some correlations exist between them.

However, our work has some limitations that inspire future directions: (1) The work is only validated on multi-modal MR brain tumor images, in the future, we will valid our method in different multi-modal image datasets. (2) The proposed correlation description block is a simple two-layer network, we intend to design a more complex and efficient correlation description block to describe the correlation between multi modalities. It will be interesting to test in future other f-divergence functions, such as Hellinger distance. (3) It will be interesting to consider other correlation expression functions to improve the segmentation performance. (4) The proposed correlation module is applied on brain tumor segmentation, we plan to apply it to synthesize additional images to cope with the limited medical image dataset or deal with the missing modality segmentation issue.

\section*{Acknowledgments}
This work was co-financed by the European Union with the European regional development fund (ERDF, 18P03390/18E01750/18P02733) and by the Haute-Normandie Regional Council via the M2SINUM project. This work was partly supported by the China Scholarship Council (CSC).

\bibliography{mybibfile}
\vspace{0.5cm}

\textbf{Tongxue Zhou} received the B.S. degree in Biomedical Engineering from Jilin University, Changchun, China in 2015. She is currently pursuing the Ph.D. degree in Computer Science from National Institute of Applied Sciences of Rouen (INSA Rouen Normandie), Rouen, France. Her current research interests include medical image analysis, data fusion and deep learning

\textbf{Su Ruan} received the M.S. and the Ph.D. degrees in Image Processing from the University of Rennes, France, in 1989 and 1993, respectively. She was a Full Professor in the University of Champagne-Ardenne, France, from 2003 to 2010. She is currently a Full Professor with the Department of Medicine, and the Leader of the QuantIF Team, LITIS Research Laboratory, University of Rouen, France. Her research interests include pattern recognition, machine learning, information fusion, and medical imaging.

\textbf{Pierre Vera} received the M.D. degree in 1993 from Université Paris VI and the PhD degree from the same institution in 1999. He is currently a University Professor and Hospital Physician with the Faculty of Medicine, University of Rouen, France. He is also the General Director of Henri Becquerel Cancer Center, and the head of the Department of Nuclear Medicine. His research interests include radiation oncology, nuclear medicine, biophysics, and medical imaging.

\textbf{Stéphane Canu} received the Ph.D. degree in system command from the Compiègne University of Technology in 1986. He received the French Habilitation degree from Paris 6 University. He is currently a Professor of the LITIS research laboratory and of the information technology department, at the National Institute of Applied Sciences of Rouen (INSA Rouen Normandie). His research interests includes deep learning, kernels machines, regularization, machine learning applied to signal processing, pattern classification and optimization for machine learning.

\end{document}